\documentclass[10pt,twocolumn,letterpaper]{article}

\usepackage{cvpr}
\usepackage{times}
\usepackage{epsfig}
\usepackage{graphicx,subfigure}
\usepackage{amsmath}
\usepackage{amssymb}
\usepackage{diagbox,booktabs}

\usepackage[breaklinks=true,bookmarks=false]{hyperref}

\cvprfinalcopy 

\def\cvprPaperID{****} 

\begin{document}

\title{\Large\textbf{Real-time Action Recognition with Dissimilarity-based Training of Specialized Module Networks}}

\author {Marian K.Y. Boktor\textsuperscript{1,4}, Ahmad Al-Kabbany\textsuperscript{2,4}, Radwa Khalil\textsuperscript{1}, and Said El-Khamy\textsuperscript{1,3}\\
\textsuperscript{1}ECE Department, Arab Academy for Science, Technology, and Maritime Transport\\
\textsuperscript{2}Intelligent Systems Lab, Arab Academy for Science, Technology, and Maritime Transport\\
\textsuperscript{3}Department of Electrical Engineering, Alexandria University\\
\textsuperscript{4}Department of Research and Development, VRapeutic\\
Alexandria, Egypt\\
alkabbany@ieee.org, elkhamy@ieee.org
}

\maketitle

\begin{abstract}
This paper addresses the problem of real-time action recognition in trimmed videos, for which deep neural networks have defined the state-of-the-art performance in the recent literature. For attaining higher recognition accuracies with efficient computations, researchers have addressed the various aspects of limitations in the recognition pipeline. This includes network architecture, the number of input streams (where additional streams augment the color information), the cost function to be optimized, in addition to others. The literature has always aimed, though, at assigning the adopted network (or networks, in case of multiple streams) the task of recognizing the whole number of action classes in the dataset at hand. We propose to train multiple specialized module networks instead. Each module is trained to recognize a subset of the action classes. Towards this goal, we present a dissimilarity-based optimized procedure for distributing the action classes over the modules, which can be trained simultaneously offline. On two standard datasets--UCF-101 and HMDB-51--the proposed method demonstrates a comparable performance, that is superior in some aspects, to the state-of-the-art, and that satisfies the real-time constraint. We achieved $72.5\%$ accuracy on the challenging HMDB-51 dataset. By assigning fewer and unalike classes to each module network, this research paves the way to benefit from light-weight architectures without compromising recognition accuracy\footnote{An earlier version of this research has appeared in \cite{EarlyVersion}}.
\end{abstract}

\section{Introduction} 
\label{sec:Introduction}
An extensive research has been conducted with the goal of developing algorithms that is capable of recognizing people'€™s actions without human intervention. The applications of such algorithms are diverse. A considerable attention has been given to military and security-related applications, such as the employment of action recognition systems at airports and border-crossings. Meanwhile, further applications have arisen in healthcare and quality-of-life enhancement, education, human-computer interfaces, in addition to others. This growing demand has stimulated highly-paced research which aims to develop new efficient and effective techniques, seeking higher recognition accuracies and lower computational requirements \cite{13}.

Deep convolutional neural networks (CNN) have defined the state-of-the-art (SOTA) performance in several classification problems, and have thus become a standard tool in the recent literature of computer vision and image/video processing research. For video applications, in particular, three-dimensional (3D) CNNs are known to be more efficient than 2D CNNs. Nevertheless, huge datasets are required for training these architectures. Large-scale datasets, such as the Kinetics dataset \cite{11}\footnote{The Kinetics dataset consists of 400 action classes with more than 400 video clips per class.}, have been proven to impact positively the performance of popular deep CNN models. Among the computer vision problems that have been tackled using CNNs is action recognition.

Action recognition is a well-studied problem in computer vision, and its literature is vast. Numerous approaches have been adopted to address this problem including Representation-based Solutions, such as Holistic Representations \cite{8}, Local Representations and Aggregation Methods \cite{14}. In relation to the scope of this research, we limit the discussion to deep networks-based approaches. One possible classification of the literature of \emph{deep action recognition} is presented in \cite{18}, where approaches are grouped based on the type of their input, namely, color videos approaches, skeleton sequences and depth maps approaches. Various network architectures have been introduced as a part of SOTA frameworks that aim to overcome the open problems in action detection and recognition. One of a special interest, which belongs to the former family of techniques, and a prominent example of such architectures, is the two-stream CNNs \cite{1}. Essentially, it processes simultaneously and benefits from fusing two types of features for the input videos. This approach increases the prediction accuracy, but it trades off computational complexity and real-time performance when the adopted features are computationally demanding (the optical flow is an example). Other researchers have proposed alternative types of features, that are computationally efficient, to augment the color information. These techniques have benefited from the effectiveness of the two-stream architecture and have attained real-time performance, at the expense of reduced accuracies. The work of \cite{7} is an example of this family, and it adopts highly informative features (motion vectors) using a transfer learning-based algorithm.

In this research, we present a novel scope for multi-stream architectures that are capable of attaining SOTA accuracies and meeting real-time requirements. Using color information only, we propose to break down the recognition task over multiple specialized module networks (streams). Each module is trained to recognize a subset of action classes in the dataset at hand. Our modules adopt an architecture that learns residual features (ResNets) \cite{9}, pre-trained on the Kinetics dataset. To facilitate the recognition, we develop a dissimilarity-based optimized procedure that places unalike actions classes together in each module network (stream). We adopt a pipeline that is ended by feeding deep features to a classical classifier. Several types of classifiers such as SVM, MLP and LDA have been adopted, and we report the results of the best performers. Lastly, we propose different approaches for fusing the decision of each stream (each classifier) at the inference time. 

\begin{figure*}[t]
	\begin{center}
		\setlength{\abovecaptionskip}{0pt plus 0pt minus 0pt}
		\setlength{\belowcaptionskip}{-15pt plus 0pt minus 0pt}
		\subfigure[Training pipeline]{\includegraphics[width=6.5in]{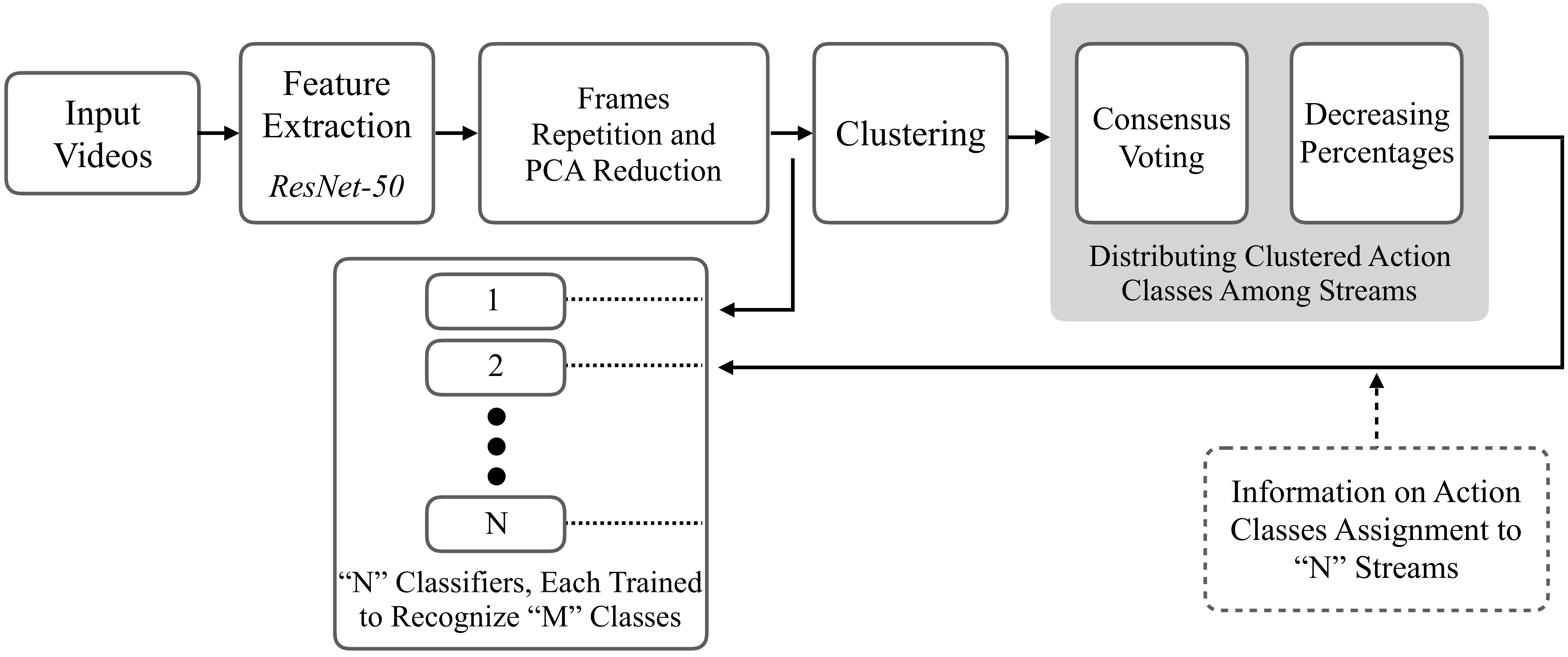}}
		
		\subfigure[Testing pipeline]{\includegraphics[width=5.5in]{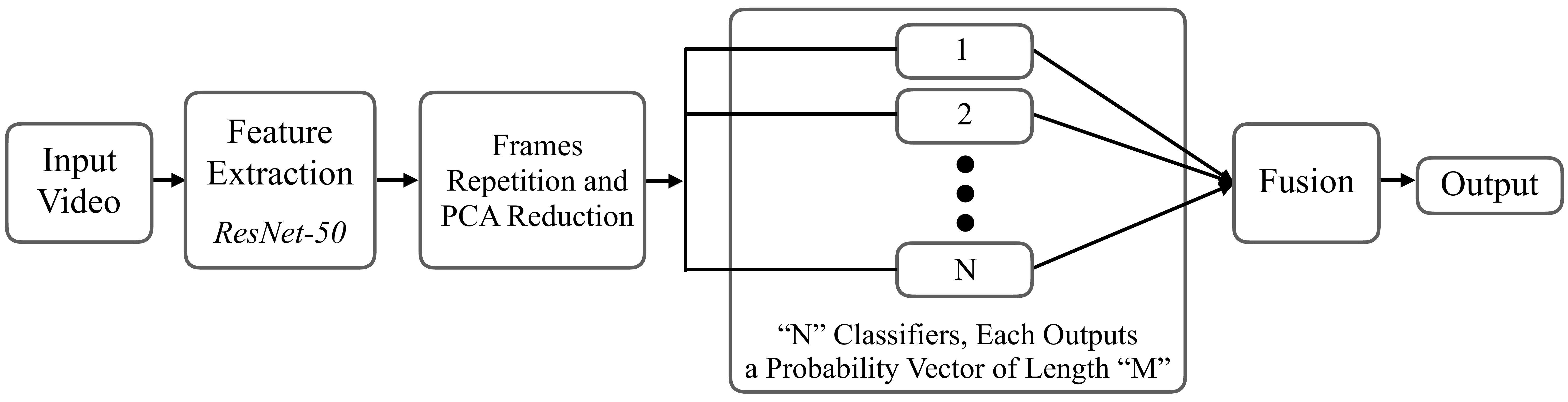}}
		\caption{The training and testing pipelines of the proposed method}
		\label{fig:Pipeline}
	\end{center}
\end{figure*}

Comparing the proposed method with SOTA techniques on two standard datasets has shown comparable performance that is superior in some aspects. This is achieved without compromising the real-time constraint. Because the action classes are distributed effectively among streams during training, placing dissimilar classes together, the proposed model paves the way towards benefiting from module networks with light-weight architectures, such as MobileNet. The proposed pipeline is shown in Fig.~\ref{fig:Pipeline}, and we elaborate on it in Sec.~\ref{sec:ProposedMethod}.

The rest of this paper is organized as follows:  Additional related work is discussed in Sec.~\ref{sec:RelatedWork}. A detailed description of the proposed framework is found in Sec.~\ref{sec:ProposedMethod}.  Results and discussions on the conducted experiments are described in Sec.~\ref{sec:Results}. Finally, conclusions and future work are presented in Sec.~\ref{sec:Conclusion}.

\section{Related Work}
\label{sec:RelatedWork}
Deep networks-based techniques are the main focus of the action recognition research community nowadays. Most of such techniques utilize Convolutional Neural Networks (CNNs) in order to train a specific model on completing the classification at hand. These deep models have attained SOTA results \cite{11, 3, 6}. 

One type of those networks, Spatio-Temporal Networks, have an architecture that helps the model to learn spatio-temporal features \cite{4}\footnote{Given  authors' instructions, \cite{4} is cited for using code on GitHub. \cite{2} is cited for the details that are not found in \cite{4}.}. Spatial features are generally important to get insights about objects' positions and their interrelations within an image, while temporal features are essential for capturing temporal information in image sequences and video, that is necessary for the activity recognition problem. 

A relatively new, and widely-used, framework in deep action recognition is Multiple Stream Networks. Two-stream 2D CNNs are considered to be the first implemented multiple stream network. It enhances the performance by fusing the information from two or more streams after assigning different roles to each stream. For instance, Simonyan et al. introduced a novel procedure \cite{1}, that uses RGB images as still frames to extract appearance information in one stream. Simultaneously, it combines color information with optical flow frames, stacked together to represent motion information across the temporal dimension in the other stream. 

Recently, different strategies employing 3D CNNs have outperformed the existing 2D CNNs approaches, such as C3D \cite{5} and 3D ResNets \cite{4}. 2D CNNs and their variants that are pre-trained on ImageNet were very successful in image recognition tasks. 3D CNNs perform much better with regards to action recognition as well \cite{11}. Lately, Kinetics--a large-scale dataset comparable to ImageNet in terms of volume--offered a remarkable improvement in recognition accuracies when used in pre-training deep networks \cite{4}. Hara et. al, explored ResNet models with a different number of layers and architectures, pre-trained on Kinetics. They achieved a top accuracy of 90.7\%  using ResNext-101 architecture on UCF-101. This is close to the accuracy of ResNet-50 (89.3\%) which we adopt in the proposed work as our feature extractor as detailed in Sec.~\ref{sec:ProposedMethod}. 

To the best of our knowledge, only a limited number of techniques are available in the literature of real-time deep action recognition. One example is the method proposed by Zhang et. al. \cite{7}. They employed a two-stream CNN architecture with multiple transfer learning techniques in order to make effective use of motion vectors, instead of optical flow. Hence, this approach improves the overall system performance by speeding up inference for a reasonable degradation in accuracy level. The main  focus of this paper is speeding up the processing time without compromising the performance.

All of the aforementioned deep networks use an end-to-end learning framework. Another framework involves the extraction of features from a CNN network, preferably a pre-trained one, and then feed those \emph{deep features} to another classifier, such as SVM \cite{5, 12}. This latter framework has demonstrated excellent results on several datasets. 

\section{Proposed Method}
\label{sec:ProposedMethod}
The pipeline of the proposed method starts with the preprocessing of all the constituent clips of the dataset at hand. We have chosen the standard UCF-101 \cite{16} and HMDB-51 datasets to evaluate our technique. Deep features are extracted from the clips of the dataset using a pre-trained ResNet-50 model. The action classes are then distributed amongst $N$ different streams. This distribution is based on the dissimilarity of the classes. When given $M$ action classes in a dataset, the $\frac{M}{N}$ classes in each stream are guaranteed (using an optimized procedure) to be farther apart from one another than from the rest of the classes in the dataset. It is worth mentioning that the proposed method differs from ensemble learning in that the latter combines different classifiers that are trained on the same classes, while the former integrates identical classifiers trained on different action classes. In our experiments, the number of streams was fixed to $4$, and generally, this is a free parameter in the pipeline. Nevertheless, as the number of streams decreases, the benefit of breaking down the recognition task (over multiple streams) also decreases. We argue that the \emph{upper bound} for the number of streams comes from the limitation on the number of parameters in the pipeline, since every additional stream implies adding another classifier to the pipeline.

\subsection{Feature Extractor:ResNet-50}
ResNet is a type of deep convolutional neural networks. It learns residual features by creating shortcut straight connections between specific layers \cite{9}. Different variants of ResNet, such as ResNet-34, ResNet-50, and ResNet-101 are commonly used in applications where feature extraction is needed. They are comprised of 34, 50,  and 101 layers respectively. In this research, we use the ResNet-50 architecture, with a publicly available pre-trained model on Kinetics, for feature extraction \cite{2}. Based on our analysis, deeper ResNet architectures do not yield significant improvements in recognition, and require more computations which slows down inference. Following feature extraction, frames repetition takes place to end up with equal-sized feature vectors. We are aware that this step could be achieved using different approaches including the adoption of other network architectures such as LSTM. The last step before clustering is the PCA dimensionality reduction, which yields the final feature vector for each video.

\subsection{Clustering}
Following feature extraction, the action classes are then divided into $\frac{M}{N}$ clusters, where $M$ is the number of classes and $N$ is the number of streams. Clustering is done by applying K-means using Lloyd'€™s algorithm. Particularly, the PCA-reduced feature vectors of the videos in the training set is fed to the clustering algorithm. The number of PCA components is set to 1000 in our experiments. This procedure introduces two main challenges, namely, problematic action classes and unequal clusters sizes. We elaborate on these challenges in the following lines. It is worth mentioning that for UCF-101, only 25\% of the clips in each action class are randomly selected and included in the clustering process, while for HMDB-51, all of the training set is included.

We observe that for some classes, the majority of the member clips (in some cases, all of the member clips) are affiliated to one specific cluster. While for other classes, their member clips are scattered among different clusters. We define \emph{a problematic action class} (PC) as a class which has more than 50\% of its affiliates belonging to different clusters. In sub-section \ref{subsec:HandleClassAffiliation}, we explain our approach for handling PCs in more details. Another observed computational issue is that the generated clusters have different sizes. This is mainly due to the scattering of the member clips of the PCs. Meanwhile, in some cases, clips from \emph{non-PCs} are also found to be randomly mis-located. This problem is treated by proposing a new strategy for class assignment to clusters named the \emph{Decreasing Percentage Method} which is also explained in sub-section \ref{subsec:HandleClassAffiliation}.

\subsection{Handling Classes Affiliation}
\label{subsec:HandleClassAffiliation}
\begin{figure}[t]
	\begin{center}
		\includegraphics[width=3in]{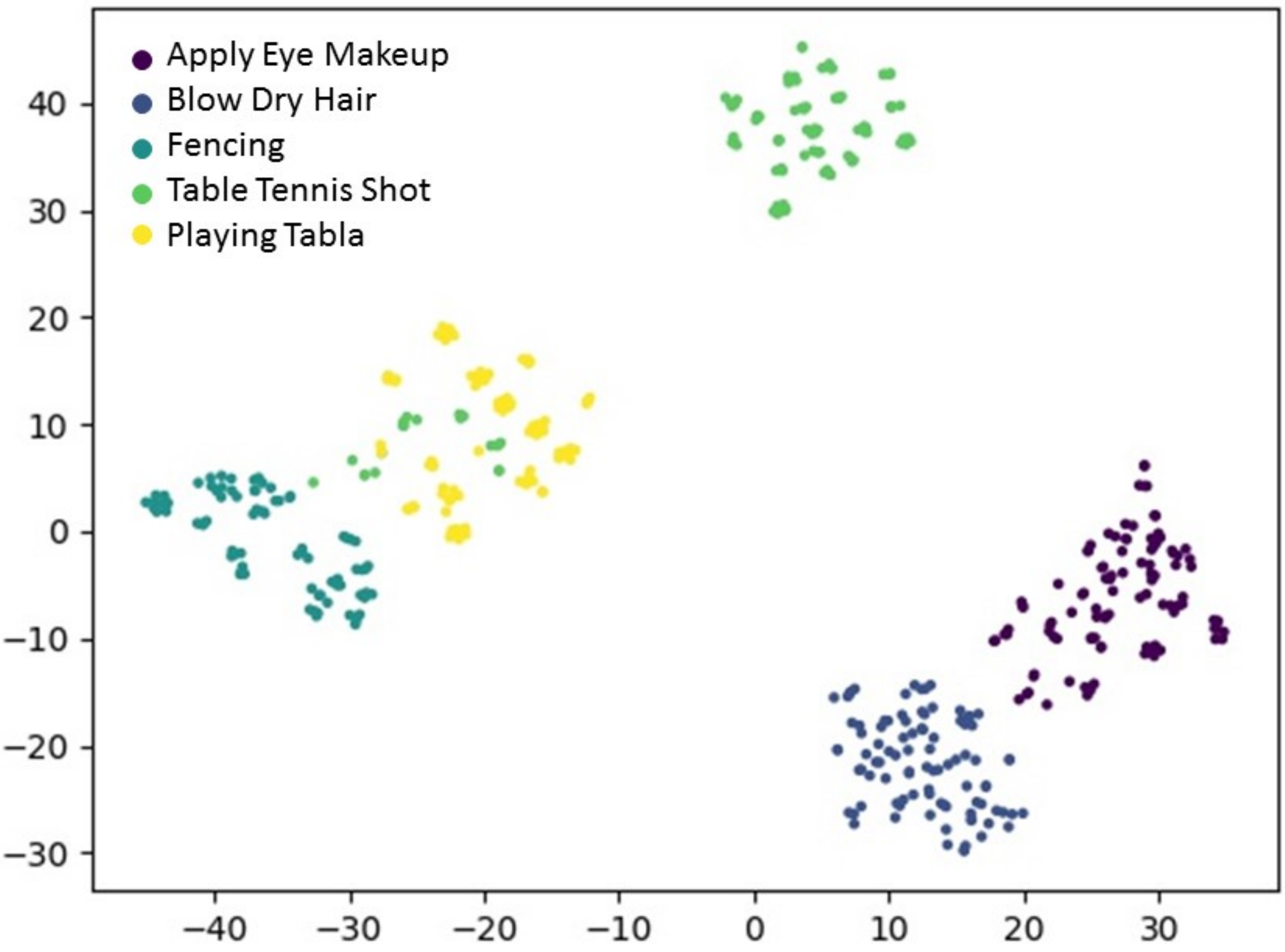}
		\caption{t-SNE plot of a PC (Table Tennis Shot) against non-PCs}
		\label{fig:ProbTabTen}
	\end{center}
\end{figure}

\textbf{Consensus Voting.} Consensus Voting is a decision-making procedure that is based on a group-level settlement approval rather than an individual'€™s preference. This method first obtains the features of all the members of each PC and finds the clusters to which those members are affiliated. For each PC, we compute the approximate nearest neighbors (ANN) for its members amongst the members of the non-PCs in those clusters. Based on the votings, we assign every PC to the cluster which includes more neighbors to its own members. Let us assume we have a problematic action class with four videos, such that {\tt{video 1}} and {\tt{video 4}} are affiliated to {\tt{cluster A}}, while {\tt{video 2}} and {\tt{video 3}} are affiliated to {\tt{cluster B}}. The non-PCs in {\tt{cluster A}} are {\tt{X}} and {\tt{Y}}, and the non-PCs in {\tt{cluster B}} are {\tt{W}} and {\tt{Z}}. The affiliation of every member video in the PC is determined based on the number of their nearest neighbors either in {\tt{X}} and {\tt{Y}}, or in {\tt{W}} and {\tt{Z}}. The member videos then vote for the \emph{collective affiliation} of their class. Members who failed to decide their own affiliation, do not participate in the action class-wide voting. Figure~\ref{fig:ProbTabTen} depicts a 2D visualization (t-SNE plot) of the members of one example of a problematic action class--\emph{Table Tennis Shot}--in the UCF-101 dataset.

\begin{figure}[h]
	\begin{center}
		\includegraphics[width=3.2in]{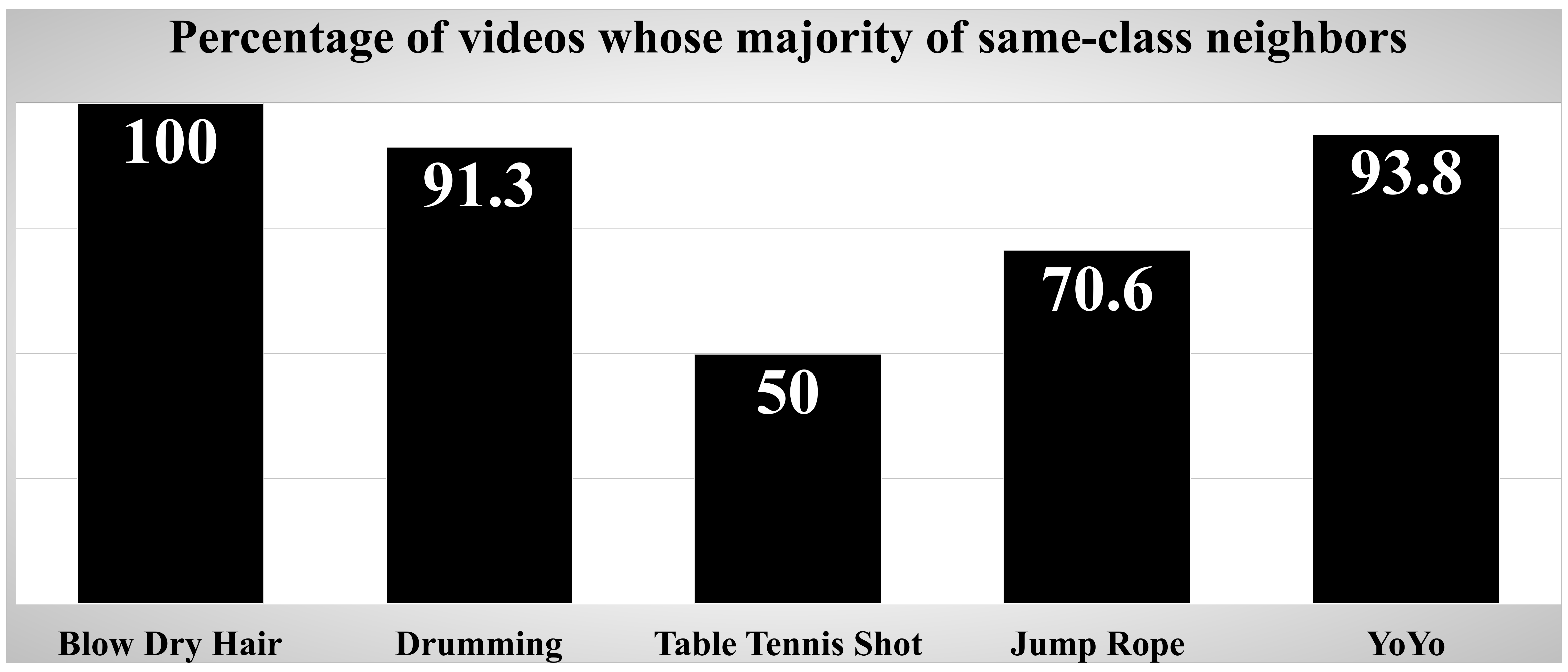}
		\caption{Analysis of five action classes of UCF-101 dataset using ANNs}
		\label{fig:ANNVerify}
	\end{center}
\end{figure}

Lastly, because we compute an ANN field in the Consensus Voting step, it is worth mentioning that we are aware of the k-nearest neighbors limitations in higher dimensions \cite{knnHigh}. In order to verify the computation of an ANN field on the datasets at hand, we chose $5$ action classes from UCF-101, including PCs, and we computed 100 neighbors for each of their members. The percentage of same-class neighbors is shown in Fig.~\ref{fig:ANNVerify}. It shows that the ANN field succeeds to capture a same-class neighbor percentage of $\geq 50\%$ consistently.

\begin{figure}[]
	\begin{center}
		\setlength{\abovecaptionskip}{5pt plus 0pt minus 0pt}
		\setlength{\belowcaptionskip}{-25pt plus 0pt minus 0pt}
		\includegraphics[width=2.7in]{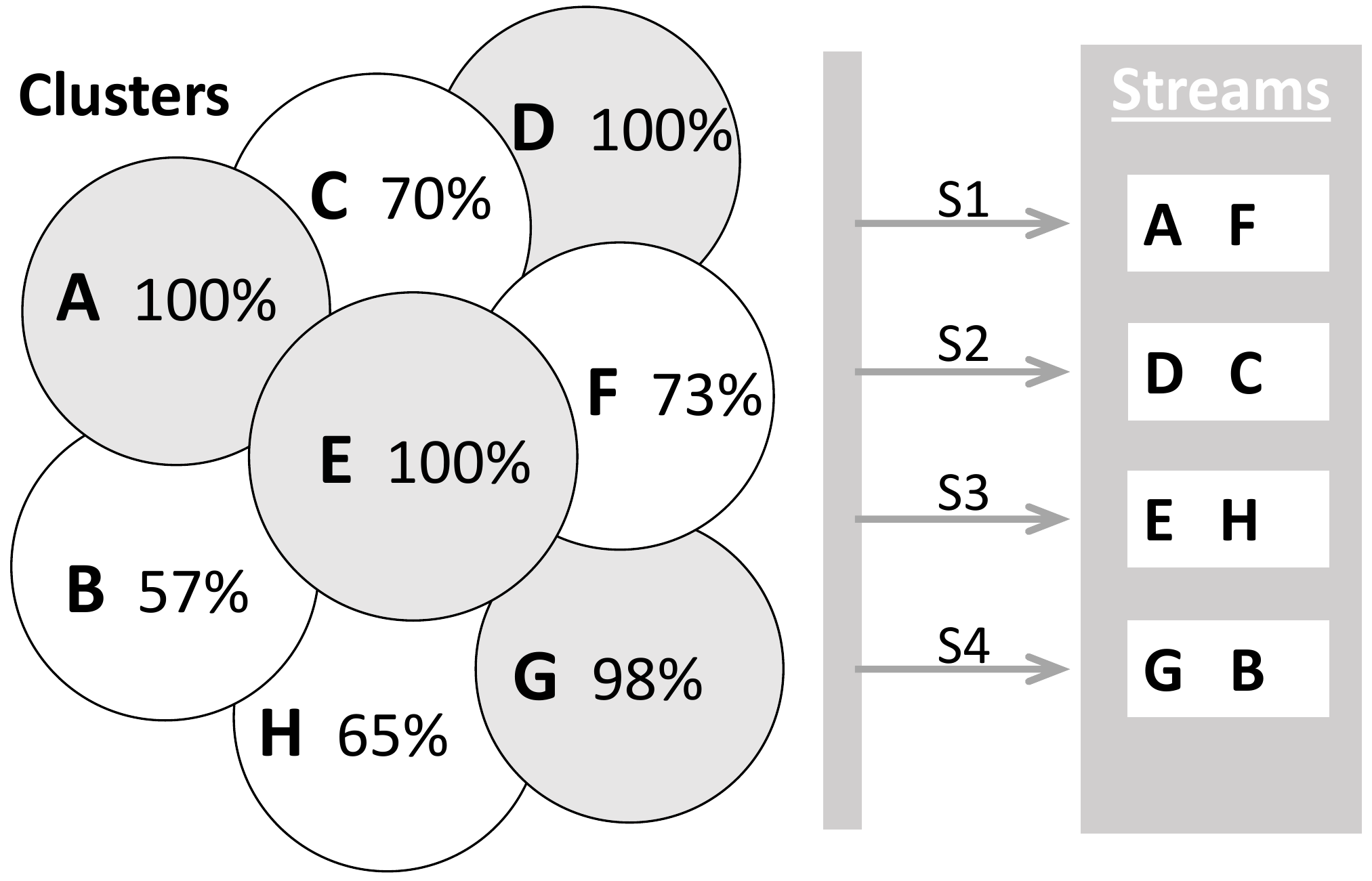}
		\caption{An illustration for Consensus Voting and Decreasing Percentages}
		\label{fig:CVDP}
	\end{center}
\end{figure}

\textbf{Decreasing Percentages method.} Consensus Voting helps matching classes to clusters. However, the main drawback of this voting technique is that, while undergoing the assignment process, it preserves the dissimilarity-based class distribution among streams without taking into consideration to avoid variable number of classes in each cluster. Given that the number of streams is set to four in our experiments, this causes a problem if the number of classes in each cluster exceeds four. To overcome this issue, the classes are arranged in each individual cluster in descending order of \emph{affiliation percentage}. The affiliation percentage of each class to its cluster refers to the percentage of particular class members in the cluster to which it is assigned. When the first four classes from each cluster are distributed amongst the four streams, the remaining classes from each cluster are then distributed in the same manner. This methodology guarantees that members with the highest affiliation percentages of a specific cluster will not be trained through the same stream. In other words, Consensus Voting helps matching action classes to clusters, while Decreasing Percentages helps matching action classes to streams. Figure~\ref{fig:CVDP} illustrates the distribution procedure. The positive impact of the CV and the DP procedures on the recognition accuracies will be discussed in Sec.~\ref{sec:Results}.

\subsection{Training and Testing}
The $N$ streams are now ready to be trained with $\frac{M}{N}$ classes in each stream (25 classes in 3 streams and 26 in one stream for UCF-101 dataset). Several types classifiers have been adopted including SVM, MLP, LSTM, and LDA. The best results were obtained from the first two types, and we elaborate on their results in the next section. PCA-reduced features of both training and testing sets from each class are fed into the stream to which they belong. First, the recognition accuracy of each stream is computed individually (without considering the output of the other streams) to confirm the robustness of the individual streams. Then these outputs are fused using various approaches which will be explained in the following sub-section. It is worth mentioning that while each stream's individual accuracy is key to reliable recognition, the approach adopted for the fusion of the decision of the different streams also heavily impacts the performance.

\subsection{Fusion}
\label{subsec:Fusion}
The most critical step in the proposed framework is the fusion of the classification outputs from the $N$ streams. Failing to orchestrate the decision of the streams can yield low recognition accuracies, even if each stream manages to correctly predict the action classes it is trained on.
The task of finding the best matching target value for each test video clip is a two-step voting process. First, voting among all the \emph{known output classes} of each stream takes place, followed by voting over the winning classes from all the $N$ streams. This means that we intervene before the classifier'€™s last stage to get the output probabilities of the $\frac{M}{N}$ classes per stream. In order to achieve high recognition accuracy, three approaches are adopted, namely, Maximum Raw Probability ($W_{raw}$), Mean-Weighted Probability ($W_{mean}$) and Minimum-Weighted Probability ($W_{min}$).

\noindent \textbf{Maximum Probability.} To start with, in every stream, we simply determine the maximum score out of the $\frac{M}{N}$ (number of classes per stream) scores and compare it to the other maximum scores of the other streams. The winner is the class with the maximum probability over all of the streams.

\noindent \textbf{Mean-Weighted Probability.} The aforementioned mechanism remarkably reduces the overall video-level accuracy to 89.7\% compared to streams'€™ average accuracy of 96.5\%. Another more sophisticated fusion mechanism is done by weighting the output probabilities before computing the maximum probability. One approach for scaling the probabilities is to use the Euclidean distance between the test clip and the mean feature vector of each class (hence the name). Then the probabilities are multiplied (weighted) by the inverse of those Euclidean distances before computing their maximum.

\noindent \textbf{Minimum-Weighted Probability.} A more computationally expensive weighting method is to evaluate the probabilities according to the minimum obtained Euclidean distance against members of each action class. This is done by computing the minimum Euclidean distance between the test clip and all of the member clips in the $N$ winning classes. The correct class should correspond to the clip with the minimum Euclidean distance. This method yields a higher video-level recognition accuracy at a high computational demand. We show a comparison between the three approaches with regards to the attained accuracies in Sec.~\ref{sec:Results}.

\section{Results and Discussion}
\label{sec:Results}
We evaluate the proposed method on UCF-101 and HMDB-51--two of the most challenging standard actions datasets. For the former, in the conducted experiments, the three train/test splits that the dataset authors provide (for a fair comparison between frameworks) are followed. As a start, the experiments are conducted on split 1 only, and then the average accuracy is calculated over the three splits in order to compare the proposed work to other SOTA techniques. The real-time-ness performance is measured in frames per seconds (fps) on an 8-core (Intel® Core i7-7700K) CPU and a single (GeForce GTX 1080) GPU. Features extraction stage is done on GPU, while all of the other stages are carried out on CPU.

The first experiment investigated the impact of the number of PCA components of the extracted features on the overall performance, using UCF-101 videos and an SVM classifier. This is done using 10, 100 and 1000 PCA components on 10, 20 and 25\% of the data, randomly chosen and split randomly in a ratio of 80 (train) to 20 (test). Results in Table \ref{table:PCACompCompare} show that using 1000 PCA components yielded slightly higher accuracies than the case of 100 components. However, because we wanted to examine the impact of the proposed method (which is the breakdown of the learning task) independently of other factors (such as throwing some information by reducing the number of considered PCA components) we preferred to retain the whole 1000 PCA components, and this was fixed throughout our experiments on UCF-101. As for HMDB-51, due to the fact that it is a relatively small sized dataset, after the clustering procedure, PCA components are set to 800 components as the number of train videos per stream ranges from 800 to about 900 videos. It is worth mentioning that for SVM classifiers, before training, the effect of different kernels and learning rates is examined on 25\% of the data with split of 80-20 (train-test) percent. The best results are achieved by RBF kernel and learning rate of $\gamma=0.0001$. Hence, the same SVM settings are used throughout the rest of the experiments.

\begin{table}[t]
	\caption{Comparison of SVM output accuracy percentages using different number of PCA components, and different data percentages that are split randomly in a ratio of 80:20 (train:test) on UCF-101. SVM settings: SVM kernel: rbf, C: 10, gamma: 0.0001}
	\label{table:PCACompCompare}
	\begin{tabular*}{8.3cm}{c||ccc}
		\toprule
		\backslashbox{Data \%}{PCA Comp.}          & 10    & 100    & \textbf{1000} \\ \hline\hline
		\addlinespace[0.25em]
		10         & 65.9 & 86.6 & \textbf{86.9} \\ 
		\addlinespace[0.25em]
		20 & 71.3 & 90.7 & \textbf{91.1}  \\ 
		\addlinespace[0.25em]
		25         & 69.5 & 90.7 & \textbf{90.9}  \\ 
		\bottomrule
	\end{tabular*}
\end{table}

Next, we compare the recognition accuracies in case of using the CV and DP procedures to the case of random assignment of classes to streams. This means that after clustering the data, both extra classes of cluster size more than $N=4$ in our case (as mentioned in sub-section \ref{subsec:HandleClassAffiliation}) and the problematic classes are distributed randomly among the streams until fully occupied. The performance of those streams using SVM is tested. Table~\ref{table:StreamsIndividualAccuracies} shows an average increase of approximately 2\% on employing the CV and DP procedures.

\begin{table}[]
	\caption{Streams individual accuracies of fine-tuning ResNet-50 on UCF-101 split 1, with and without applying Consensus Voting (CV) and Decreasing Percentages (DP) procedures}
	\label{table:StreamsIndividualAccuracies}
	\begin{tabular*}{8.3cm}{c||cccc}
		\toprule
		\backslashbox{METHOD}{STREAM}          & 1    & 2    & 3    & 4    \\ \hline\hline
		\addlinespace[0.25em]
		RANDOMLY ASSIGNED         & 97.2 & 94.9 & 95.1 & 93.0 \\ 
		\addlinespace[0.25em]
		AFTER CV and DP & 98.6 & 96.3 & 97.5 & 96.0 \\ 
		\bottomrule
	\end{tabular*}
\end{table}

At the fusion stage, three different computing methodologies are followed as previously discussed in sub-section \ref{subsec:Fusion}. Table~\ref{table:WeightProbCompare} highlights the significant increase in accuracies when employing Euclidean distances as a weighting factor. It is worth mentioning that the consistency between clip and video levels accuracies shows how the used framework is robust and reliable, in addition to being real-time. 

\begin{table}[]
	\caption{Accuracy percentages of our method using SVM on UCF-101 split-1 for the three methods of weighing probability vectors}
	\label{table:WeightProbCompare}
	\begin{center}
		\begin{tabular*}{6.5cm}{c||cc}
			\toprule
			\textbf{Method} & 	\textbf{Clip-level} & \textbf{Video-level}  \\
			& \textbf{Accuracy} & \textbf{Accuracy} \\ \hline\hline
			\addlinespace[0.25em]
			$W_{raw}$         & 90.0 & 89.7  \\ 
			\addlinespace[0.25em]
			$W_{mean}$ & 93.6 & 93.5   \\ 
			\addlinespace[0.25em]
			$W_{min}$         & \textbf{94.2} & \textbf{93.9}  \\ 
			\bottomrule
		\end{tabular*}
	\end{center}
\end{table}

Using $W_{mean}$ and $W_{min}$, Table~\ref{table:ComparisonSelf} shows the recognition accuracies of the proposed method using the two best performing classifiers from the set that included SVM, MLP, LDA, and LSTM. Also, $W_{mean}$ and $W_{min}$ video-level accuracies are calculated and averaged over three splits then compared to other SOTA results, against their speed in (fps). The proposed method achieved the highest accuracy ($94.0\%$ and $72.5\%$ on UCF-101 and HMDB-51, respectively) at 216.9 fps. It is still considered to be a real-time system since the requirements for full HD videos is only 60 fps. Whereas an end-to-end learning using ResNet-50 attains an accuracy of 89.3\% \cite{4}, the proposed method outperforms that approach on the same dataset using the same model (ResNet-50).

\begin{table}[]
	\caption{Three splits accuracy percentages for UCF-101 and HMDB-51 datasets and their average. Computed using SVM and MLP classifiers with the two weighing probabilities methods (clip-level accuracies are given in brackets in case of UCF-101)}
	\begin{center}
		\setlength{\abovecaptionskip}{-20pt plus 0pt minus 0pt}
		\setlength{\belowcaptionskip}{-25pt plus 0pt minus 0pt}
		\includegraphics[width=3.3in]{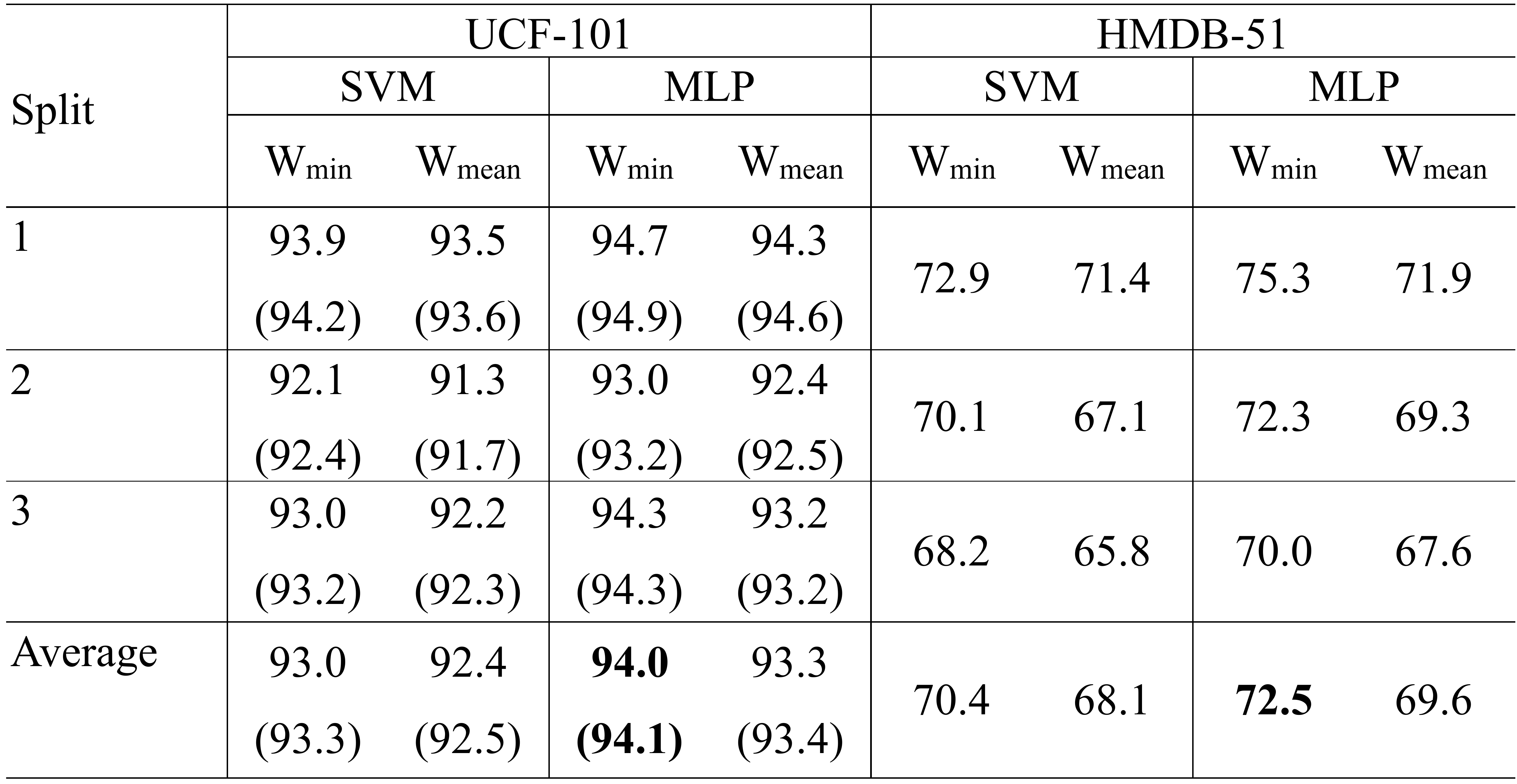}
	\label{table:ComparisonSelf}
	\end{center}
\end{table}

\begin{table}[]
	\caption{Performance of our method on UCF-101 and HMDB-51 averaged over 3-splits using SVM and MLP classifiers with the two weighing probabilities methods in comparison with other real-time frameworks}
	\begin{center}
		\setlength{\abovecaptionskip}{-20pt plus 0pt minus 0pt}
		\setlength{\belowcaptionskip}{-25pt plus 0pt minus 0pt}
		\includegraphics[width=3.3in]{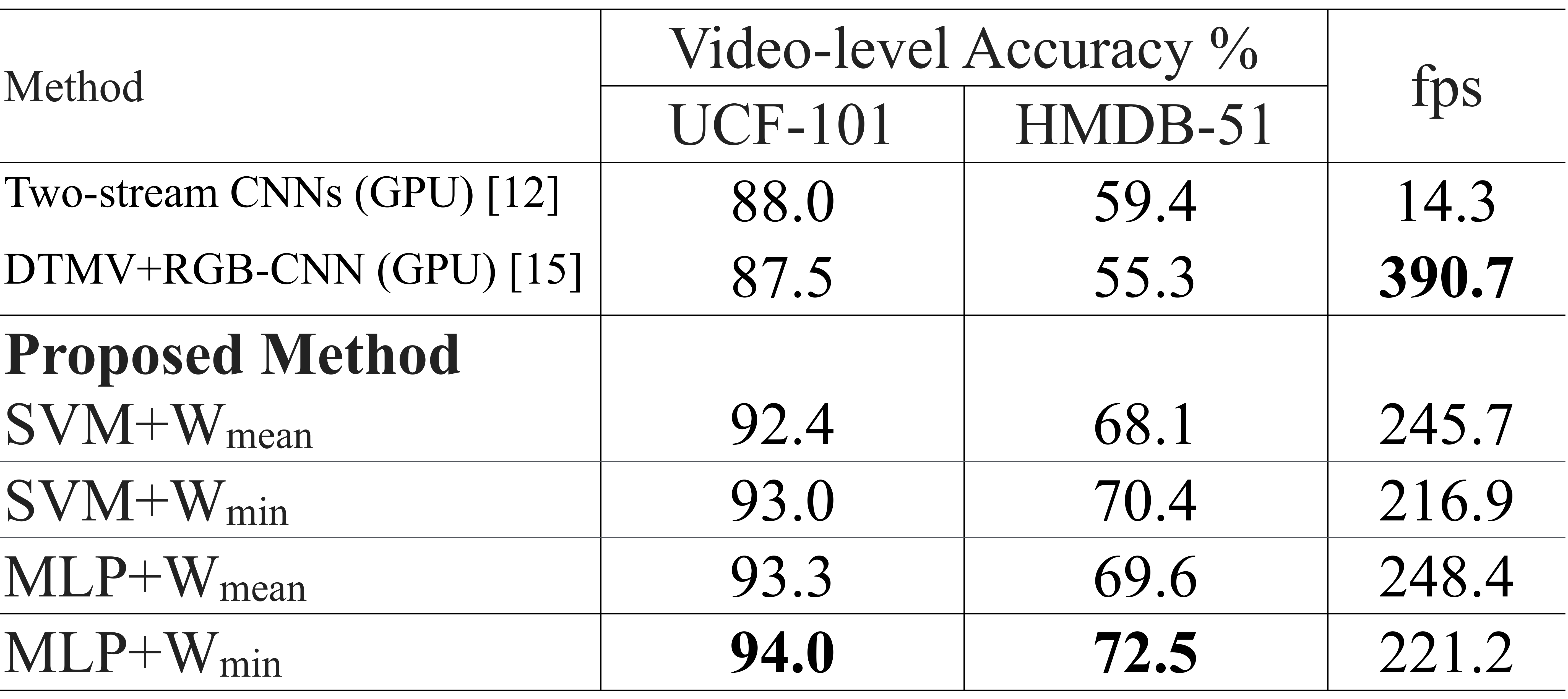}
	\label{table:ComparisonGPU}
	\end{center}
\end{table}

\noindent \textbf{Speed Calculations.} The frames per second (fps) measurements are calculated by dividing the total number of frames in the test videos by the total time taken for on-line processing. On-line processing starts from feature extraction, all the way through data preparation until producing the output score. On-line processes include feature extraction, segment number equalization, PCA computations, individual classifiers prediction and output fusion. It was not overlooked that the feature extraction part takes more than $85-95\%$--depending on the fusion method--of the total testing time, which is a major attribute that we have considered this stage of the pipeline for enhancements in our future work.

\section{Conclusion}
\label{sec:Conclusion}
This research proposed a novel framework for real-time action recognition in trimmed videos. Different from the recent literature of deep learning-based models, we proposed to break down the recognition task over multiple specialized module networks, each of which is taught to recognize a subset of action classes in the dataset at hand. We designed and evaluated an optimized procedure for assigning unalike actions classes to each module network (stream). Afterwards, we proposed different procedures for fusing the decision of each module at the inference time. Rather than adopting an end-to-end deep learning approach, we followed the recent approach of feeding deep features to a classical classifier which improves the accuracy significantly. We identified the free parameters in our proposed model and presented the analysis based on which their values were set. A comparison with SOTA methods was conducted on two standard datasets, and our proposed model has shown comparable performance that is superior in some aspects, without compromising the real-time constraint. We argue that our proposed model paves the way towards adopting module networks with light-weight architectures, thanks to our procedure for assigning fewer and unalike action classes to each module. Future work involves the investigation of more efficient fusion approaches, further investigation for the free parameters in the model, the adoption of MobileNet architecture \cite{17}, and the generalization of the specialized module networks to other classification problems.

\section*{Acknowledgement}
The authors would like to express their deep gratitude to Dr.Hassan Eldib, of the Intelligent Systems Lab at the Arab Academy for Science and Technology, for the invaluable support,  time, and help that he has given while providing and setting up the equipment that was used for carrying out the simulations in this research. The authors are also grateful for Dr.Ibrahim Sobh, of the self-driving cars research team at Valeo Egypt for the insightful discussions which enhanced the quality of presentation of this article.

\bibliographystyle{ieee}
\bibliography{myRef}


\end{document}